\documentclass[11pt]{article}
\usepackage[utf8]{inputenc}
\usepackage{times}

\usepackage[numbers]{natbib}
\usepackage{url}
\usepackage{rotating}
\usepackage{epsfig} 
\usepackage{graphicx}
\usepackage{subfigure}
\usepackage{fancyhdr}
\usepackage{parskip}
\usepackage{enumitem}
\usepackage{txfonts}
\usepackage{times}
\usepackage{multirow}

\usepackage{floatflt, lscape, graphicx}
\usepackage{wrapfig}
  \graphicspath{{figs/}}
  \DeclareGraphicsExtensions{.pdf,.jpg,.png,.eps}  
\usepackage{latexsym}
\usepackage[ruled]{algorithm}
\usepackage[noend]{algorithmic}
\usepackage{xspace}
\usepackage{longtable}
\usepackage[pdfstartview=FitV,bookmarks=false]{hyperref}
\urldef\myhomepage\url{https://users.cecs.anu.edu.au/~hannakur}
\usepackage{titlesec}
\usepackage{blindtext}

\newcommand{\ie}{\textrm{i.e.}}

\newcommand{\cref}[1]{~\cite{#1}}

\newcommand{\sref}{Section~\ref}

\newcommand{\aref}[1]{Algorithm~\ref{#1}}
\newcommand{\eref}[1]{eq.~(\ref{#1})}

\newcommand{\fref}[1]{Figure~\ref{#1}}
\newcommand{\tref}[1]{Table~\ref{#1}}

\newenvironment{myEnumerate}
{\begin{enumerate}[noitemsep, topsep=-8pt, parsep=0pt, partopsep=0pt]
}
{\end{enumerate}\vspace{2pt}}

\newenvironment{myDescription}
{\begin{description}[noitemsep, topsep=-6pt, parsep=0pt, partopsep=0pt]
}
{\end{description}\vspace{2pt}}

\newcounter{comment}

\newcounter{secCtr}
\addtocounter{secCtr}{0}
\newcounter{subSecCtr}
\addtocounter{subSecCtr}{1}
\newcounter{tSecCtr}
\addtocounter{tSecCtr}{1}

\newcommand{\mdpTuple}{\textrm{$\langle {\mathcal{S}}_{MDP}, {\mathcal{A}}_{MDP}, T_{MDP}, R_{MDP}\rangle$}\xspace}
\newcommand{\mdpStSpace}{\textrm{${\mathcal{S}}_{MDP}$}\xspace}
\newcommand{\mdpActSpace}{\textrm{${\mathcal{A}}_{MDP}$}\xspace}
\newcommand{\mdpTrans}{\textrm{$T_{MDP}$}\xspace}
\newcommand{\mdpRew}{\textrm{$R_{MDP}$}\xspace}

\newcommand{\pomdpTuple}{\textrm{$\langle \mathcal{S}, {\mathcal{A}}, \mathcal{O}, T, Z, R\rangle$}\xspace}

\newcommand{\ub}[1]{\textrm{$\overline{V}(#1)$}\xspace}
\newcommand{\qub}[1]{\textrm{$\overline{Q}(#1)$}\xspace}
\newcommand{\lb}[1]{\textrm{$\underline{V}(#1)$}\xspace}
\newcommand{\qlb}[1]{\textrm{$\underline{Q}(#1)$}\xspace}

\newcommand{\belSpace}{\textrm{$\mathcal{B}$}\xspace}
\newcommand{\belTree}{\textrm{$\mathcal{T}$}\xspace}
\newcommand{\belSet}{\textrm{$B$}\xspace}
\newcommand{\belInit}{\textrm{$b_0$}\xspace}
\newcommand{\bel}{\textrm{$b$}\xspace}
\newcommand{\belp}{\textrm{$b'$}\xspace}
\newcommand{\belpp}{\textrm{$b''$}\xspace}

\newcommand{\belc}{\textrm{$b_c$}\xspace}

\newcommand{\genMod}{\textrm{${\mathcal{G}}$}\xspace}
\newcommand{\hist}{\textrm{$h$}\xspace}
\newcommand{\histp}{\textrm{$h'$}\xspace}

\newcommand{\stSpace}{\textrm{$\mathcal{S}$}\xspace}
\newcommand{\st}{\textrm{$s$}\xspace}
\newcommand{\stp}{\textrm{$s'$}\xspace}
\newcommand{\stc}{\textrm{$s_c$}\xspace}
\newcommand{\goalSt}{\textrm{$G$}\xspace}

\newcommand{\actSpace}{\textrm{${\mathcal{A}}$}\xspace}
\newcommand{\act}{\textrm{$a$}\xspace}
\newcommand{\actp}{\textrm{$a'$}\xspace}

\newcommand{\obsSpace}{\textrm{$\mathcal{O}$}\xspace}
\newcommand{\obs}{\textrm{$o$}\xspace}

\newcommand{\transFunc}{\textrm{$T$}\xspace}
\newcommand{\transProb}{\textrm{$P(s' \,|\, s, a)$}\xspace}
\newcommand{\errS}{\textrm{$\eta$}\xspace}
\newcommand{\errDist}{\textrm{${\mathcal{N}}$}\xspace}
\newcommand{\transBel}{\textrm{$\tau$}\xspace}

\newcommand{\obsFunc}{\textrm{$Z$}\xspace}
\newcommand{\obsProb}{\textrm{$P(o \,|\, s', a)$}\xspace}

\newcommand{\rewFunc}{\textrm{$R$}\xspace}
\newcommand{\rew}{\textrm{$r$}\xspace}

\newcommand{\polOpt}{\textrm{$\pi^*$}\xspace}
\newcommand{\pol}{\textrm{$\pi$}\xspace}
\newcommand{\alpPolTree}{\textrm{$T_{\pi_{\alpha}}$}\xspace}

\newcommand{\alpSet}{\textrm{$\Gamma$}\xspace}
\newcommand{\alp}{\textrm{$\alpha$}\xspace}
\newcommand{\reach}{\textrm{${\mathcal{R}}$}\xspace}
\newcommand{\reachOpt}{\textrm{${\mathcal{R}}^*$}\xspace}
\newcommand{\ubSet}{\textrm{$U$}\xspace}
\newcommand{\ubPt}{\textrm{$u$}\xspace}

\begin{document}


\title{Partially Observable Markov Decision Processes  (POMDPs) and Robotics}

\author{Hanna Kurniawati \\
School of Computing, Australian National University\\
hanna.kurniawati@anu.edu.au}

\date{}

\maketitle

\begin{abstract}
Planning under uncertainty is critical to robotics. The Partially Observable Markov Decision Process (POMDP) is a mathematical framework for such planning problems. It is powerful due to its careful quantification of the non-deterministic effects of actions and partial observability of the states. But precisely because of this, POMDP is notorious for its high computational complexity and deemed impractical for robotics. However, since early 2000, POMDPs solving capabilities have advanced tremendously, thanks to sampling-based approximate solvers. Although these solvers do not generate the optimal solution, they can compute good POMDP solutions that significantly improve the robustness of robotics systems within reasonable computational resources, thereby making POMDPs practical for many realistic robotics problems. This paper presents a review of POMDPs, emphasizing computational issues that have hindered its practicality in robotics and ideas in sampling-based solvers that have alleviated such difficulties, together with lessons learned from applying POMDPs to physical robots. 
\end{abstract}



\section{Introduction}

The ability to compute reliable and robust decisions in the presence of uncertainty is essential in robotics. Specifically, an autonomous robot must decide how to act strategically to accomplish its tasks, despite being subject to various types of errors and disturbances affecting their actuators, sensors, and perception, and despite the lack of information and understanding about itself and its environment. The errors and limited information cause the effects of performing actions to be non-deterministic from the robot's point of view and cause the robot's state to only be partially observable, which means the robot never knows its exact state.  

The Partially Observable Markov Decision Process (POMDP)\cref{drake1962observation, sondik1971optimal} is a  mathematically principled framework to model decision-making problems in the non-deterministic and partially observable scenarios mentioned above. The POMDP quantifies the non-deterministic effects of actions and errors in sensors and perception stochastically. It estimates the robot's state as probability distribution functions over states, called beliefs, and computes the best actions to perform with respect to these beliefs, rather than single states. The computed action strategies will automatically balance information gathering and goal attainment. This concept is powerful: It is general and could enable robust operation even when the robot operates near environment boundary or near the limit of the robot's capability. 

However, exactly because of its careful consideration of uncertainty, computing the exact optimal solution to a POMDP problem is computationally intractable\cref{papadimitriou1987complexity}. In fact, not long ago, most benchmark problems for POMDPs have less than 30 states and the best algorithms that could solve them took many hours\cref{kaelbling1998planning, monahan1982state}, which is grossly insufficient for realistic robotics problems. As a result, POMDPs were considered impractical for robotics and abandoned at the expense of robustness.

Nevertheless, in the past two decades, tremendous advances have been made in computing good action  strategies for POMDP problems, thanks to the sampling-based approach. Although the computed strategies are not the optimal solution to the problems, they are often sufficient to substantially improve robustness. Hence, these progress enable the POMDP to become practical for a variety of realistic robotics problems.

In this paper, we describe an overview of these advances in POMDPs. We start by describing the POMDP problems and model in \sref{s:problemDef}. Subsequently, we describe sampling-based methods that have advanced the practicality of POMDPs in robotics and the computational issues these methods alleviated. In \sref{s:implementation}, we present the implementation side of POMDPs in relation to robotics applications. Finally, we end with a brief discussion on the similarity of the progression of POMDPs and motion planning, as well as the relation between POMDPs and machine learning.

\section{The Problem and POMDP Formulation}
\label{s:problemDef}

The POMDP is a natural representation of sequential decision-making problems where the results of actions are non-deterministic and the state is only partially observable. Sequential decision-making (aka. planning) is the problem of computing action strategies for a robot to achieve good long-term returns when actions may have long-term consequences. In such  problems, the robot has some information about the effects of actions prior to execution, though these information may not be perfect nor complete. In other words, the robot's understanding of the actions' results are non-deterministic. The robot can use the perceived observations to help infer its state. However, in partially observable scenarios, due to errors in sensor measurements and in perception, the robot may perceive the same observations from multiple states, causing these states to be indistinguishable and the robot's exact state to remain unknown.  

Many robotics problems fit the above planning in non-deterministic and partially observable scenarios. For example:  
\begin{myDescription}
\item[Underwater Navigation:]  How should an Autonomous Underwater Vehicle (AUV) navigates to a pre-specified goal, despite not knowing the exact underwater currents affecting its motion and despite substantial localization errors underwater?
\item[Manipulation:] How should a robot pick up  an oil container from one location to another when it does not know exactly how full the container is? This lack of information means a relevant property of the problem is partially observable and the robot has uncertainty on the effect of its grasping. For example, if the container is almost empty, it will be easily moved and perhaps fall over when the robot tries to pick it up from the side.
\item[Human Robot Collaboration:] How to communicate effectively, so as to ensure effective collaboration with human, even though the robot does not know the exact characteristics nor intentions of the human? These variables are partially observed and due to a lack of information about the human characteristics and intention, the reaction of the human with respect to the robot's actions becomes non-deterministic.
\end{myDescription}
The above examples are obviously far from being exhaustive in the robotics topics nor in the problems within each robotics topic, but hopefully they gave an indication of how diverse and common non-deterministic and partially observed planning problems are in robotics. 

The above type of planning problems can naturally be formulated as a POMDP. Formally, the POMDP model is defined as a 6-tuple \pomdpTuple, where:
\begin{myDescription}
	\item[\stSpace] denotes the state space ---the set of all possible {\em  states}, which can include the states of the robot and the environment.
	\item[\actSpace] denotes the action space ---the set of all {\em actions} the robot can perform.
	\item[\obsSpace] denotes the observation space ---the set of all {\em observations} the robot can perceived. 
	\item[$\transFunc(s, a, s')$] denotes the transition function, representing the non-deterministic effects of actions. It is a conditional probability function \transProb representing the probability that the robot will be in state $\stp \in \stSpace$ after performing action $\act \in \actSpace$ at state $\st \in \stSpace$. In robotics, this function is sometimes represented as a noisy dynamics function $\stp = f(\st, \act, 
	\errS)$, where $\st, \stp \in \stSpace \subseteq \mathbb{R}^n$ and $\errS \sim \errDist$ is a noise vector sampled from noise distribution \errDist, while $f(\cdot)$ denotes the system's dynamics.
	\item[$\obsFunc(s', a, o)$] denotes the observation function, representing errors and noise in measurement and perception. It is a conditional probability function \obsProb that represents the observation the robot may perceive when it is in state $\stp \in \stSpace$ after performing action $\act \in \actSpace$. 
	\item[\rewFunc] denotes the immediate reward function. This function can be parameterized by a state, a pair of state and action, or a tuple of state, action, and subsequent state.
\end{myDescription}\vspace{6pt}

\begin{figure}[!htbp]
	\centering
	\vspace{-0.1cm}
	\includegraphics[width=12cm]{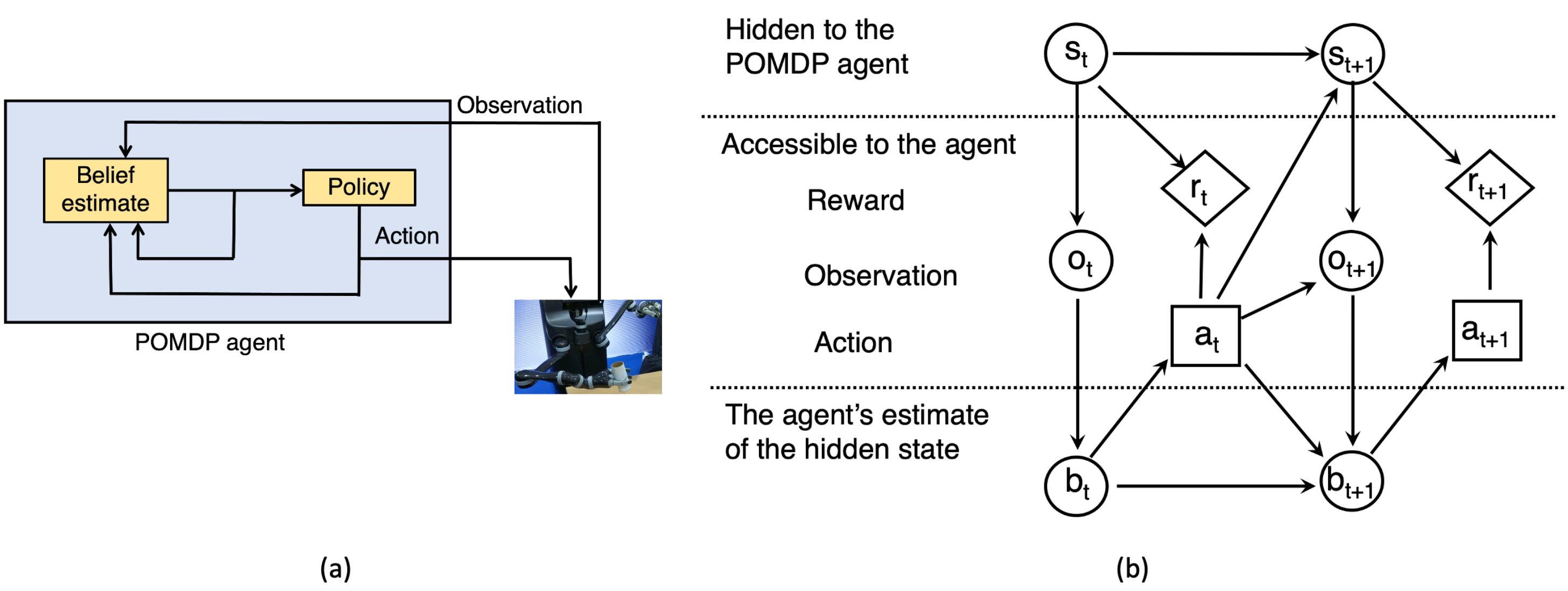}\vspace{0cm}
	\caption{Illustration of a single time-step of a POMDP agent (a) and process (b).\vspace{0cm}}
	\label{f:pomdpIllustrate}
\end{figure}

A POMDP agent with model \pomdpTuple will operate as follows. At each time step, the agent is in some state $\st \in \stSpace$. However, due to partial observability, \st is hidden to the agent, and instead the agent maintains a belief $\bel \in \belSpace$ as an estimate of its state, with the notation \belSpace denoting the belief space (i.e., the set of all beliefs). The agent infers the \emph{best} action $\act \in \actSpace$ to execute from \bel (what \emph{best} means is defined in the following paragraph). Once the action is performed, the hidden state may move to a new state $\stp \in \stSpace$. The state \stp is hidden to the agent, but the agent perceives an observation $\obs \in \obsSpace$ that may reveal some information about \stp. The possible state \stp the agent moves to and the observation \obs it may perceive follows the transition  \transFunc and observation functions \obsFunc, respectively. Since the state \stp is hidden to the agent, the agent updates its estimate of the state from \bel to belief \belp via Bayesian inference based on the previous estimate \bel, the action \act it just performed, and the observation \obs it just perceived. Finally, the agent receives a reward, based on the reward function \rewFunc, from which the objective function, and hence best action is derived from. This sequence forms a single step of a POMDP agent, and the process repeats. \fref{f:pomdpIllustrate} illustrates this single step. 

Solving a POMDP problem modelled as \pomdpTuple  means finding an optimal policy ---that is, a mapping $\polOpt: \belSpace \rightarrow \actSpace$ from beliefs to actions that maximise the objective function. Many objective functions have been proposed. One that is often used in robotics is the following value function, which is based on the expected total discounted reward. This value function assumes the problem has an infinite horizon, meaning, at each time step, the POMDP agent can still move infinitely many steps. 
\begin{equation}
	V^*(b) = \max_{\act \in \actSpace} \underbrace{\left(R(b, a) + \gamma \underbrace{\sum_{\obs \in \obsSpace} P(o \,|\, b, a) V^{*}(\transBel(\bel, \act, \obs))}_{\textstyle{J(\bel, \act)}} \right)}_{\textstyle{Q(\bel, \act)}}
\label{e:optValFunc}
\end{equation}
where $R(b, a) = \sum_{\st \in \stSpace} R(s, a) \cdot b(s)$ is the expected immediate reward, while $\transBel(\bel, \act, \obs)$ updates the belief estimate \bel after the agent performs action $\act \in \actSpace$ and perceives observation $\obs \in \obsSpace$. Suppose $\belp = \transBel(\bel, \act, \obs)$, then
\begin{eqnarray}
\belp(\stp) = P(\stp \,|\, \obs, \act, \bel) &=& \frac{P(\obs \,|\, \stp, \act, \bel) \, P(\stp \,|\, \act, \bel)}{P(\obs \,|\, \act, \bel)} \nonumber \\ 
&=& \frac{\obsFunc(\stp, \act, \obs) \, \sum_{\stp \in \stSpace} \, \transFunc(\st, \act, \stp) \, \bel(\st)}{\sum_{s'' \in \stSpace}\obsFunc(s'', \act, \obs) \, \sum_{\st \in \stSpace} \transFunc(\st, \act, s'') \, \bel(\st)}  
\label{e:bel}
\end{eqnarray}
The probability $P(\obs \,|\, \act, \bel)$ can be computed as a normalizing factor, \ie, the denominator in \eref{e:bel}, to ensure the belief \belp sums to one. The notation $\gamma \in (0, 1)$ is a discount factor to ensure that the objective function for infinite horizon problems is well defined. Finding the best action from a belief \bel then involves solving an optimization of the Q-value $Q(\bel, \act)$ for \bel and computing an estimation of the expected future total reward $J(\bel, \act)$.


A related objective function is the finite horizon, where the POMDP agent has a finite number of steps to perform. In this case, the value function is non-stationary, and the expected total reward for a problem with horizon $T$ is $V_T^*$, as defined below:
\begin{equation}
	V_t^*(b) = \max_{\act \in \actSpace} \left( R(b, a) + \gamma \sum_{\obs \in \obsSpace} P(o \,|\, b, a) V_{t-1}^{*}(\transBel(\bel, \act, \obs)) \right) \;\; \textrm{where }\;
	V_0^*(b) =  \max_{\act \in \actSpace} \left( R(b, a) \right) 
	\label{e:finiteHor}
\end{equation}
where $V_t^*(b)$ is the value of \bel when the POMDP agent can still move for $t$ steps.

Another related and commonly used objective function in robotics is Goal-POMDP, or otherwise known as Shortest Path POMDP. Goal-POMDP assumes the state space contains a set of goal states. Let's denote this set of goals as $\goalSt \subset \stSpace$. The reward function of a Goal-POMDP problem reflects the cost of actions, and the objective is then to reach a target belief with the lowest total cost. A target belief \bel is one where $\bel(s) = 0$ whenever $s \not\in \goalSt$. A Goal-POMDP is equivalent to a POMDP with expected total discounted reward\cref{bonet2009solving}, in the sense that one can be transformed into another without changing the optimal policy and value function.

Throughout this paper, we will focus on POMDPs with expected total discounted reward (\eref{e:optValFunc}). The optimal value function of such POMDPs can be approximated arbitrarily closely by a Piecewise Linear Convex function\cref{sondik1978optimal}. Furthermore, this infinite horizon objective function has a benefit that the optimal value function, and hence the optimal policy, is stationary. 

Note that the state, action, and observation spaces of a POMDP model can be discrete or continuous. When the state and/or observation spaces are continuous, the summation in \eref{e:optValFunc} -- \eref{e:finiteHor} are replaced with integrations over the respective spaces. In this paper, we focus on discrete and finite state, action, and observation spaces, unless otherwise stated.

\section{Sampling-Based  Approximate POMDP Solvers} 
\label{s:pomdpSolvers}

Finding the optimal solution to a POMDP problem is PSPACE-hard\cref{papadimitriou1987complexity}. Different sub-classes of POMDPs have slightly different hardness results, though most are still hard for classes  above P\cref{mundhenk2000complexity, vlassis2012computational}. Note, however, that planning under uncertainty in robotics is known to be a hard problem. For instance, motion planning for a 3D point robot with uncertainty in control and localization is PSPACE-hard\cref{natarajan1986moving}, and if this robot is compliant, in the sense that when the robot is commanded to move through an obstacle, it complies with the obstacles' geometry instead of forcing itself to go through an obstacle\cref{canny1987new}, the problem is NEXP-hard\cref{canny1987new}. These results indicate that in general, the robotics problems of planning in non-deterministic and partially observable scenarios are computationally hard, even if they are not formulated as  POMDPs.

Many methods to find the optimal policy to POMDP problems have been proposed. A survey of such methods are available in \cref{monahan1982state}. However, the high computational complexity of these methods made them impractical for many realistic robotics problems. 

A major breakthrough for POMDPs' applications  in robotics comes when the sampling-based approximate POMDP solver\cref{Pin03:Point} demonstrate that it can compute good policies for a problem with 870 states, in contrast to problems with under 30 states, which was the majority of the benchmark at the time. In this paper, we focus on the sampling-based approach for computing good POMDP policies and describe details of some of the methods under this approach in \sref{s:largeStSp}--\sref{s:complexDyn}. Now, let's first discuss an overview of the approach.

Sampling-based approximate POMDP solvers relax the optimality requirement to approximate optimality and restricts the problem only to scenarios where the POMDP agent starts from a given initial belief (let's denote this belief as \belInit). Key to the approach is it samples a set of representative beliefs and computes the best action to perform only from the set of sampled beliefs, rather than the entire beliefs, thereby substantially reducing the complexity of finding good POMDP policies. Which set would be sufficiently representative and how difficult would it be to find such a set have been explored in \cref{Hsu07:What}, utilising the notion of set cover.

Many methods under the above mentioned approach have been proposed. However, they  can in general be abstracted into the program skeleton in \aref{a:samplingPOMDP}. 
\begin{algorithm}[!htbp]
\caption{A typical program skeleton for sampling-based POMDP solvers}
\label{a:samplingPOMDP}
\begin{algorithmic}[1]
	\STATE Initialize policy \pol and a set of sampled beliefs \belSet \\
	\COMMENT{Generally, \belSet is initialised to contain only a single belief (e.g., the initial belief \belInit)}
	\REPEAT
	\STATE Sample a (set of) beliefs 
	\COMMENT{Some methods sample histories (a history is a sequence of action--observation tuples) rather than beliefs. In POMDPs, beliefs provide sufficient statistics of the entire history\cref{hauskrecht2000value}, and therefore the two provide equivalent information}  
	\STATE Estimate the values of the sampled beliefs \\
	\COMMENT{Generally, via a combination of heuristics and update / backup operation}
	\STATE Update \pol
	\COMMENT{In most methods, this step is a byproduct of the previous step}
	\UNTIL{Stopping criteria is satisfied}
\end{algorithmic}
\end{algorithm}
These methods iteratively sample a set of beliefs and estimate the values of these sampled beliefs. A variety of sampling strategies have been proposed and are often critical to the performance of the method. 
Similarly, multiple methods have been proposed to estimate the values of the sampled beliefs. Some methods use heuristics to sample the newly sampled beliefs and backup operations to propagate these new information to improve the estimated values of other sampled beliefs.  
In general, this backup automatically updates the policy \pol. The process is repeated until a stopping criteria is satisfied. 

Most sampling-based approximate POMDP solvers are anytime, which means they can return a solution when stopped at any time, though of course there is a trade-off between the quality of the solution and the time the method has run. Therefore, the stopping criteria for this approach to solving POMDPs is often set to be the available planning time. Some methods\cref{Smi04:Heuristic, Smi05:Point, Kur08:Sarsop} compute upper and lower bound estimates of the value functions, and hence the stopping criteria can be set to be when the difference between the upper and lower bounds for the initial belief \belInit is less than a pre-specified threshold. However, for practical purposes, for most realistic robotics problems, even methods that compute these upper and lower bounds often stop before the desired threshold gap is reached. 

Many sampling-based approximate POMDP solvers can be broadly divided into offline and online. Offline solvers compute an approximately optimal POMDP policy \pol prior to execution. During execution, the agent only needs to estimate its current belief and execute the action $\pol(\bel)$. Online solver, on the other hand, interleaves policy computation and execution: At each step, prior to execution, the solver will compute the good action $\act \in \actSpace$ to perform from the current belief \bel. Once the action \act is performed, the agent perceives an observation, updates its belief, and the process repeats.  

Regardless of offline or online, these sampling-based methods aim to alleviate one or more of the following difficulties, which is crucial to enable POMDPs to become practical in robotics. 
\begin{myEnumerate}
\item Large state space
\item Long planning horizon
\item Large observation space 
\item Large action space 
\item Complex transition dynamics
\end{myEnumerate} 
Among these issues, large state space and long planning horizon are two most discussed issues to date, often referred to as the curse of dimensionality and the curse of history, respectively. However, other issues become equally important when applying POMDPs to realistic robotics problems. In fact, the difficulty of solving a POMDP problem is influenced by a combination of problem characteristics related to the above issues, together with additional problem characteristics, such as the sparsity of the transition and observation functions.  The work in \cref{Hsu07:What} derives a criteria that captures these combined problem characteristics to identify how difficult different POMDP problems are for sampling-based methods, though this derived criteria is not always easy to compute. 

In the next subsections, we describe the above issues in finding good POMDP policies, together with the sampling-based methods that have been proposed to explicitly alleviate them.  Although the methods presented are not exhaustive, but we hope they provide some insights on the ideas that have significantly improve the practicality of POMDPs in robotics.

\subsection{Large State Space}
\label{s:largeStSp}

This issue is known as the curse of dimensionality: A POMDP solver must reason in the belief space, which is an $(n-1)$ dimensional continuous space, where $n$ is the size of the state space. 
Key to sampling-based methods is that they restrict estimating the value function only on a set of representative beliefs, selected in an inexpensive manner via sampling.  They relax the optimality requirement to substantially improve the scalability of POMDP solvers, with large state space being one of the first issues these solvers try to address. Below are some of the offline and online sampling-based methods that directly try to address the issue of large state space.

\textbf{Offline Methods}

Point-Based Value Iteration (PBVI)\cref{Pin03:Point} was the first approximate POMDP solver that demonstrated good performance on problems with hundreds of states, i.e., an 870 states Tag (target finding) problem, albeit taking $\sim$50 hours. 

The \alp-vectors representation is used as the policy representation of PBVI. This representation maintains a finite set of \alp-vectors, denoted as \alpSet, and assumes the state space \stSpace is finite and represents beliefs as discrete probability vectors. Recall that the optimal value function \eref{e:optValFunc} can be approximated arbitrarily closely by a Piecewise Linear Convex function. Therefore, it can be rewritten as $V^*(\bel) = \max_{\alp \in \alpSet}\; \alp \cdot \bel$, where 
$\alp \cdot \bel$ represents the inner product between the two vectors, whose sizes are the same as the number of states in \stSpace. Intuitively, each $\alp \in \alpSet$ corresponds to a policy tree \alpPolTree, where each node is associated with an action in \actSpace and each edge is associated with an observation in \obsSpace. The value $\alp(\st)$  is then the expected total reward of starting from state $\st \in \stSpace$, executing the action associated with the root of \alpPolTree, traversing down the path in \alpPolTree based on the observation perceived, and executing the associated actions at each node of \alpPolTree. Each $\alp \in \alpSet$ corresponds to an action $\act \in \actSpace$, which is the action associated with the root of \alpPolTree. When a vector $\alp \in \alpSet$ maximizes $V^*(\bel)$, the policy  maps the belief \bel to the action $\act \in \actSpace$ that is associated with \alp. More details about \alp-vectors representation of a POMDP policy are available in\cref{kaelbling1998planning}. 
 
PBVI samples beliefs from the set \reach(\belInit) of beliefs reachable from a given initial belief \belInit that are far from the already sampled beliefs.  Specifically, given a set of sampled beliefs $\belSet \subset \belSpace$, PBVI expands the set by performing a single-step forward simulation for each pair of belief $\bel \in \belSet$ and action $\act \in \actSpace$. It then keeps only one of the resulting beliefs for each $\bel \in \belSet$ as the newly sampled belief and add it to \belSet. The belief kept is the one farthest away from any belief already in \belSet based on L1 metric.

Point-based backup is used by PBVI to estimate the value function (\eref{e:optValFunc}). This backup operation computes the value function (\eref{e:optValFunc}) only at a finite set of sampled beliefs. It maintains a single \alp-vector for each sampled belief. Given an existing policy \alpSet and a newly sampled belief \bel, the new vector \alp that corresponds to \bel is constructed by assigning $\alp_a(s) = R(s, a)$ and  computing $\alp = \arg\max_{a \in {\mathcal{A}}} \; \alp_{b, a} \cdot \bel$, where $\alp_{b,a} = \alp_a + \sum_{o \in O} \arg\max_{\alpha \in \Gamma}\; \alpha \cdot  \transBel(\bel, \act, \obs)$. Finally, \alp is added to \alpSet.

The idea of applying backup operation on only a finite set of beliefs have been proposed since the early work on POMDPs\cref{smallwood1973optimal} and multiple subsequent works\cref{cheng1988algorithms, kaelbling1998planning}. However, to ensure optimality, these methods select the set of beliefs systematically, which is expensive. The work in \cref{zhang2001speeding} introduces Point-based Dynamic Programming Update with backup operation that is very similar to the backup operation of PBVI. It selects beliefs based on some heuristics, which is much faster than the systematic selection proposed in \cref{smallwood1973optimal, cheng1988algorithms, kaelbling1998planning}.  However, they interleave  point-based backup with the much more expensive standard dynamic programming backup, to ensure optimality of the solution. By relaxing the optimality requirements, PBVI performs only point-based backup and replaces the  expensive belief selection with inexpensive belief sampling, resulting in significant scaling up of POMDP solving capabilities.

Another sampling-based method, Perseus\cref{spaan2005perseus}, separates belief sampling from backup operation, in the sense that backup is not performed to all sampled beliefs. Perseus uses \alp-vectors to represent the value function and policy and uses point-based backup too. However, by performing backup operation only on a subset of the sampled beliefs, it generates a smaller set of \alp-vectors,  and hence reduces the memory requirements. 

Later offline methods substantially improve the performance of PBVI and Persues further. For instance, Heuristic Search Value Iteration (HSVI)\cref{Smi04:Heuristic}, and specifically HSVI2\cref{Smi05:Point}, took 2 hours to generate a policy for Tag that has better quality than the policy generated by PBVI after 50 hours. Whilst, Successive Approximations of the Reachable Space under Optimal Policies (SARSOP)\cref{Kur08:Sarsop} generates a better policy for Tag than the one generated by HSVI2 with only 6 seconds computation time. Since then, HSVI2 and SARSOP have been demonstrated to generate good policies for problems with over 15K states and 1K observations, while SARSOP has also been shown to generate good policies  for RockSample(10,10) benchmark\cref{Smi04:Heuristic}, which has over 100K states\cref{ong2010planning}. Below, we present an overview of both HSVI2 and SARSOP, highlighting their strategies for sampling beliefs. 

HSVI2 uses \alp-vectors policy representation and point-based backup, but differ from PBVI in its sampling strategy. HSVI2 maintains a lower and upper bound estimates of the value function, where the upper bound is used to guide sampling and is initialized with the value function of the fully observable (i.e., the Markov Decision Process (MDP)) simplification of the POMDP problem. This upper bound is represented as a set of points $\ubSet \subset \belSpace \times \mathbb{R}$ and computed using sawtooth approximation\cref{hauskrecht2000value}. The lower bound is the current policy and is represented as a set \alpSet of \alp-vectors. Each sampled belief $\bel \in \belSet$ is associated with a lower and upper bound, denoted as \lb{\bel} and \ub{\bel}, where \lb{\bel} is associated with a vector $\alp \in \alpSet$ and \ub{\bel} is associated with a point $\ubPt \in \ubSet$. 

HSVI2 maintains the set of sampled beliefs in a belief tree, denoted as \belTree, where the nodes represent beliefs and an edge labelled with a pair of action--observation \act--\obs  from \bel to \belp means there is an action $\act \in \actSpace$ and an observation $\obs \in \obsSpace$, such that $\belp = \transBel(\bel, \act, \obs)$. We will use the same notation for a node and the belief it represents. The root of \belTree represents the initial belief \belInit. HSVI2 interleaves belief sampling and backup until the gap between the upper and lower bound of \belInit is sufficiently small ---that is, $|\ub{\belInit} - \lb{\belInit}| \leq \epsilon$ for a small threshold $\epsilon$. To sample beliefs, HSVI2 performs multiple sequences of forward simulations, starting from \belInit and following a path down the tree \belTree.  Given a belief \bel, a single-step forward simulation selects the action with the best upper bound $\act = \arg\max_{\act \in \actSpace}\qub{\bel, \act}$ and observation with the highest excess uncertainty $\obs = \arg\max_{\obs \in \obsSpace} |\ub{\belInit} - \lb{\belInit}| - \gamma \epsilon^{-t} $, where $\gamma$ is the discount factor and $t$ is the depth of node \belp in \belTree. The belief $\belp = \transBel(\bel, \act, \obs)$ is then set as the child node of \bel in \belTree via an edge labelled \act--\obs. This single-step forward simulation process is repeated down the tree until the gap between the upper and lower bound of the newly added belief contribute less than the threshold $\epsilon$ to such a gap at \belInit.

Now, SARSOP uses \alp-vectors policy representation, point-based backup, maintains upper and lower bounds estimates, and  represents the set of sampled beliefs \belSet as a belief tree \belTree too. However, SARSOP explicitly aims to sample from the set of beliefs  $\reachOpt(\belInit)$ reachable from \belInit under the optimal policy. Although sampling from the set of beliefs \reach reachable from \belInit (as is PBVI and HSVI2) has significantly improved the scalability of POMDP solving, sampling useful beliefs ---that is beliefs in or around $\reachOpt(\belInit)$--- becomes increasingly harder for deeper levels of the belief tree because the size of \reach increases much faster than that of \reachOpt.  
 
Of course, \reachOpt is not known a priori, as otherwise we would have found the optimal policy. Therefore, SARSOP interleaves predicting the optimal value function with belief sampling. The  prediction step uses a simple learning mechanism, where the belief space is discretized into bins based on features of the beliefs (in this case, the initial upper bound and entropy). The predicted value of a new belief \bel in \belTree is then the average of the values of the sampled beliefs that lie in the same bin as \bel. If the bin is empty, the predicted value is set to be the upper bound. If the predicted value of \bel is higher than a target value, which indicates a better estimate of $V(\bel)$ may improve $V(\belInit)$, SARSOP proceeds to expand \bel using single-step forward simulation similar to the one used in HSVI2. The target value at \bel is the lower bound of the root \lb{\belInit} that has been propagated down from \belInit to \bel in the tree \belTree. 

Furthermore, since value estimate and belief sampling are interleaved, beliefs in \belSet that have been sampled early in the process may be based on a poor estimate of the value function. To keep \belSet small and as close as possible to the optimally reachable set $\reachOpt(\belInit)$, SARSOP prunes branches of \belTree that are provably sub-optimal ---that is, when $\qub{\bel, \act} < \qlb{\bel, \actp}$ for a node \bel in \belTree and $\act, \actp \in \actSpace$, all descendants of \bel via the edges labelled \act--*, where * is any observation $\obs \in \obsSpace$, are pruned. Furthermore, SARSOP prunes a vector $\alp \in \alpSet$ whenever it is $\delta$-dominated by another vector in \alpSet at all points in \belSet. The vector \alp is $\delta$-dominated at belief $\bel \in \belSet$ whenever $\alp \cdot \belp < \alp' \cdot \belp$ for all beliefs $\belp \in \belSet$ that are $\delta$ distance from \bel. 

We hope the relatively detailed description of the three solvers above illustrates how substantial improvement in the scalability of POMDP solving can be achieved  by altering only how beliefs are sampled, indicating the importance of this component.

The solvers described above relies on Value Iteration. Sampling-based approach has also been applied to Policy Iteration quite early on in Point-Based Policy Iteration (PBPI)\cref{ji2007point}. This methods represents policy explicitly as  a Finite State Controller (FSC) together with the value function, as represented by the set of \alp-vectors. PBPI replaces the exact policy improvement step of the Policy Iteration method in \cref{Han98:Solving} with the point-based backup used in PBVI together with PBVI's belief sampling strategy. 

All of the above solvers assume that the state space, and also the action and observation spaces, are finite and discrete. Work have been proposed to extend them to continuous spaces. Many work in this extension focus on the policy representation. For instance, \cref{thrun1999monte} proposes a point representation. Here, the value function is represented by a set of beliefs \belSet along with their estimated Q-values. Given a new belief \bel, the Q-value for \bel and each action in \actSpace can be computed as an average of the Q-values for the particular action at \bel's $k$-nearest beliefs in \belSet, where distance is computed using KL-divergence.  The method in \cref{Porta06:Point} extends point-based methods to find good policies for POMDPs with continuous state space by replacing the value function representation \alp-vectors with \alp-functions, and using Gaussian mixture to represent the belief, transition, and observation functions. Monte Carlo Value Iteration (MCVI)\cref{bai2010monte} proposes a policy graph representation, where each node $v$ in the policy graph represents an action and is associated with an \alp-function, where $\alp_v(\st)$ is the expected total reward of executing the policy graph when the agent starts at state \st and execution starts by executing the action at node $v$. Another line of work, Guided Cluster Sampling (GCS)\cref{Kur12:Global}, represents the policy using either point representation\cref{thrun1999monte} or policy graph\cref{bai2010monte}, but focuses on the belief sampling strategy to alleviate the difficulty of sampling representative beliefs when the state space of the POMDP problem has many continuous state variables. We will see in the next subsection that online methods representation that no longer requires global value function representation, such as \alp-functions, makes it easier to scale-up solving capabilities to problems with continuous state spaces.

\textbf{Online Methods}

Online methods further improve the scalability of computing good POMDP policies by focusing to compute only the best action to perform from the current belief, rather than a policy for $\reach(\belInit)$ or $\reachOpt(\belInit)$. The best action to perform is computed right before execution, and therefore time to compute them is in general very limited. However, by focusing on only the current belief, online methods have much lower memory requirements compared to offline methods, which is a major hindrance for further scalability of offline methods. 

RTDP-Bel\cref{bonet1998solving} is one of the first online sampling-based methods for solving POMDPs approximately. It is designed for Goal-POMDP. However, the method presented in\cref{bonet2009solving} can transform any discounted POMDP to Goal-POMDP. RTDP-Bel maintains a hashtable of discretized estimated values of the beliefs. Given the current belief $\bel \in \belSpace$,  RTDP-Bel performs a one-step forward simulation for each pair of \bel--\act, where $\act \in \actSpace$, and uses a heuristics to estimate the expected total future reward. Specifically, for each \bel--\act pair, it samples a state \st from \bel, a subsequent state \stp based on \transFunc(\st, \act, \stp), and an observation \obs based on \obsFunc(\stp, \act, \obs). It then computes $Q(\bel, \act) = \rewFunc(\bel, \act) + \sum_{\obs \in \obsSpace} P(o \,|\, \bel, \act)V(\belp)$ where $\belp = \transBel(\bel, \act, \obs)$. The value $V(\belp)$ is computed using a heuristics or based on the values of the beliefs within the same bin as \belp  in the hashtable, if the bin is not empty. Finally, RTDP-Bel selects the action $\actp = \arg\max_{\act \in \actSpace} Q(\bel, \act)$ to execute and updates $V(b) = Q(\bel, \actp)$ and the hashtable of estimated value of sampled beliefs. The work in \cref{bonet2009solving} demonstrated that RTDP-Bel is comparable to PBVI and HSVI2 for larger benchmark problems, such as Tag and Rock-Sample(7,8)\cref{Smi04:Heuristic}. A recent work\cref{kim2019pomhdp} have extended this method to use particle representation and multiple heuristics to guide sampling, and further demonstrate the capability of this line of work.

Another major line of work in online methods adopt the forward search idea of RTDP-Bel, but uses the Monte Carlo Tree Search (MCTS) mechanism, which reduces reliance on heuristics. Furthermore, most online methods introduced below and subsequently rely on particle representation of beliefs and particle filter to update the beliefs, thereby making these solvers scalable for POMDPs with very large and even continuous state spaces. 

The Partially Observable Monte Carlo (POMCP)\cref{silver2010monte} extends the Monte Carlo Tree Search (MCTS), and specifically the Upper Confidence bounds for Trees (UCT)\cref{kocsis2006bandit} to partially observable domain. UCT applies a multi-arm bandit method, called Upper Confidence Bound (UCB)\cref{auer2002finite}, for action selection in MCTS. POMCP does not require an explicit transition, observation, and reward functions, rather it uses a generative model $\genMod(\st, \act)$, which maps a pair of state $\st \in \stSpace$ and action $\act \in \actSpace$ to a tuple $(\stp, \obs, \rew)$, where $\stp \sim \transFunc(\st, \act, S')$, $\obs \sim \obsFunc(\stp, \act, O)$, and \rew is the reward for performing \act from \st . 

POMCP maintains a tree \belTree, where the root node corresponds to the current belief \belc. Each node of \belTree represents a belief $\bel(\hist)$ associated with its history (denoted as \hist), which is the sequence of action--observation pairs $\hist = (a_0,o_0, a_1, o_1, \cdots, a_k, o_k)$ where $\bel = \transBel( \cdots (\transBel(\transBel(\belc, a_0, o_0), a_1, o_1)\cdots), a_k, o_k)$.  
The history associated with the root node is an empty sequence.  
Furthermore, each node maintains statistical information to help guide future sampling. 
We will refer to nodes of \belTree and the beliefs associated with them interchangeably. The belief $\bel(\hist)$ is represented as a set of particles. Note, however that the belief update is only performed during execution, and not during planning, as detailed below. 

To find the best action to perform from the current belief \belc, POMCP constructs the tree \belTree with \belc as the root node and performs many forward simulations from the root node. To perform a forward simulation form \belc, POMCP samples a state \stc from \belc and uses the sampled state to guide sampling. To sample subsequent beliefs, given a node $\bel(\hist)$, that is associated with history \hist, and a state $\st \in support(\bel(\hist))$, POMCP performs a forward simulation from  $\bel(\hist)$ by selecting an action \act based on UCB1\cref{auer2002finite}, i.e., $\act = \arg\max_{\act \in \actSpace} V(\hist \act) + c \sqrt{\frac{N\left(\hist\right)}{N\left(\hist \act\right)}}$ where $N(\hist)$ is the number of times the node has been visited, $N(\hist, \act)$ is the number of times the action \act has been applied to node \bel, and $c$ is a constant to balance exploitation and exploration. The value $V(\hist\act)$ is an estimate of $Q(\bel(\hist), \act)$, which is computed as an average of the total discounted reward of multiple forward simulations. Now, let $(\stp, \obs, \rew) = \genMod(\st, \act)$, then $\histp = append(\hist, (\act, \obs))$ and \stp is added to the particles set that represents the belief $\belp(\histp)$ associated with \histp. If \histp has been visited before, the above forward simulation process is repeated from the node $\belp(\histp)$ and sampled state \stp. Otherwise, a new child of \bel is formed to correspond to the belief $\belp(\histp)$ and history \histp. The value  $V(\histp)$ of this new leaf node is estimated by computing the total discounted reward following a pre-defined policy, often called as the rollout policy. The rollout policy can be replaced with a heuristic to estimate $V(\histp)$. A good estimate or rollout policy can help compensate for a lack of scalability in the planning horizon. Forward simulation is then restarted from the root node. Once the planning time for the step is over, the best action \act from \belc is executed, an observation \obs is perceived, and the robot's belief is updated to $\belpp = \transBel(\belc, \act, \obs)$ via particle filter. The tree \belTree is reset, \belpp is set as the root node of \belTree, and the process repeats.
 
Another method, the Adaptive Belief Tree (ABT)\cref{Kur13:An}, uses MCTS similar to POMCP too, but modifies POMCP in two areas. First, ABT performs backup operation along a path of \belTree from a leaf node to the root, after each forward simulation down the tree is concluded (i.e., a new leaf node is added). This backup operation helps improve the estimated value function used for action selection in subsequent forward simulations. Second, ABT reuses previously built trees and estimated values of nearby beliefs to help improve estimating value function of newly added tree. These two modifications help ABT to generate good strategies much faster than POMCP when good actions from nearby beliefs are similar, which is common in robotics problems\cref{Kur13:An, Hoe19:Multilevel}. 

Another method is Determinized Sparse Partially Observable Tree (DESPOT)\cref{somani2013despot}. DESPOT is based on tree search and Monte Carlo sampling too, but uses a fixed number of scenarios (say $K$) to sample the beliefs. DESPOT expands every action, but uses the fixed number of scenarios to sample the observations during forward simulation. This strategy generates a sparsely sampled belief tree, with $(|\actSpace|^DK)$ many nodes for a constant $K$ for any depth $D$ of the tree. 

\subsection{Long Planning Horizon}
\label{s:longHorizon}

This issue is often known as the curse of history. To compute good long-term return, a POMDP solver performs lookahead for $k$ number of steps to consider future consequences of its action selection. However, in general, the size of the set of possible future consequences increases exponentially with the number of lookahead steps $k$.

Most methods discussed in \sref{s:largeStSp} also claim to have alleviated the problem of long planning horizon. This is true because by estimating value function only for a small subset of the reachable space $\reach(\belInit)$ and even reachable space under an optimal policy $\reachOpt(\belInit)$,  computational resources can be reallocated to perform longer look-ahead, which in turn alleviate the long planning horizon issues. 

However, the above strategies are often not sufficient for many robotics problems, where the required look-ahead can easily be 30 steps and more. Many methods have been proposed to directly alleviate these issues. They generally construct a more abstract action, and sample the belief space using this abstract action rather than the primitive single-step action. As a result, they reduce the effective planning horizon of the problem. Different methods to construct abstract actions have been proposed. Most\cref{theocharous2003approximate, he2010puma, lim2011monte} use macro-actions --that is, temporally extended sequences of actions where actions or sub-policy are run until some termination conditions are met. For example, the work in \cref{lim2011monte} constructs Partially Observable Semi-Markov Decision Processes (POSMDPs) to achieve sub-goals, and use these policies as macro-actions to offline solvers. Whilst, the work in \cref{he2010puma} proposes macro-actions to reach subgoals in online solvers. These methods require sub-goals or termination conditions to be hand-designed to generate good problem decomposition. 

Obviously, automatic generation of sub-problems are preferred. The work in\cref{Kur11:Motion} develops such an automatic generation mechanism, but  for open loop policies for sampling beliefs, rather than macro actions. It sample milestones in the state space, biasing sampling towards states with  high reward and high probability of generating useful observations. Sequences of actions to move from one milestone to another, assuming deterministic actions, become the actions used to guide sampling in the belief space. The optimality of the POMDP policy found depends on the density of the state and belief space sampling. Another method\cref{agha2014firm} restricts beliefs to be Gaussian and uses LQG as macro-actions for an extension of the Probabilistic Roadmap\cref{kavraki1996probabilistic} to belief space.  
Recently, \cref{flaspohler2020belief} successfully constructs macro-actions for general POMDP solving automatically, based on the value of information. Specifically, it constructs macro-actions from sequences of open-loop policies with low value of information. It provides bounded regret on the quality of the policy generated by these macro actions.

\subsection{Large Observation Space}

Robotics problems often have rich and large (or even continuous) observation spaces, such as a combination of laser readings, joint torques reading, RGB images, etc.. Naive uniform discretization of such an observation space often results in too fine or too coarse a discretization. Overly fine discretization causes an unnecessarily large observation space, which slows down computation of good policies, while overly coarse discretization results in an effective observation space that cannot differentiate observations that induce different decisions.  

Work have been proposed to better discretize continuous  observation space\cref{hoey2005solving} based on features derived from the value function and the associated best actions. Another work\cref{bai2014integrated} proposes an extension of the policy graph representation\cref{bai2010monte} and combines it with a classification mechanism based on estimates of the value functions to identify which observations can be grouped together. Both of these methods are designed for offline methods.

For online methods, POMCPOW\cref{sunberg2018online} uses the Double  Progressive Widening to incrementally increase the set of observations to be considered. This strategy essentially discretizes the observation space incrementally based on the sampled observations. Although built on top of POMCP, POMCPOW diverges slightly, in the sense that it requires the use of weighted particles and an explicit observation function, rather than the generative model alone. 

A more recent online method to alleviate the issue of continuous observation space is Lazy Belief Extraction for Continuous Observation POMDPs (LABECOP)\cref{Hoe21:An}, which avoids any form of  discretization of the observation space. It maintains a set of sampled episodes, which is sequences of state--action--observation--reward quadruples, but postpone belief assignment until execution. During execution, LABECOP reweights the episodes to infer a belief based on the perceived observation, the action it just performed, and its current belief estimate, using a mechanism akin to particle filter. Finally, it estimates the Q-values of the actions based on the weighted average discounted total reward of appropriate components of the episodes.

\subsection{Large Action Space}

The solvers discussed above have significantly increased the scalability of POMDPs. However, most of them can only perform well for problems with a small discrete action space (i.e., $|\actSpace| \leq 100$). To find the best action to perform, a POMDP solver must solve an optimization problem (\eref{e:optValFunc}) while estimating the Q-values of the actions, which in itself is expensive to compute. Sampling has been used to improve the speed of estimating Q-values, but  most of the above solvers finds the best action naively by enumerating all actions. As a result, finding good POMDP policies becomes prohibitively expensive for problems with continuous or large discrete action space. 

Perseus\cref{spaan2005perseus} is an offline sampling-based approximate POMDP solvers that have been extended to problems with continuous action space. It replaces maximization over all actions with sampled max operator, where maximization is computed over a random subset of the action space. GCS\cref{Kur12:Global} is another offline solver for continuous action space. It performs maximization over only a subset of the action space too, but it uses geometric information from the robot operating environment to generate sequences of action space where optimization will be performed. 
The idea of sampling actions to alleviate problems with continuous action space have been proposed for tree-based solvers too in \cref{mansley2011sample}, albeit applied to the fully observable POMDP, i.e., MDP problems. 

For POMDPs, an early work that extended tree search based solvers to problems with continuous action space is GPS-ABT\cref{Sei15:An}. Due to the cost of estimating Q-values, and not to mention their gradient, GPS-ABT proposes to use the simplest non-gradient based optimization method, Generalized Pattern Search. The work also proposes an efficient data structure to efficiently keep track and reuse partially estimated Q-values of different pairs of belief--action. Despite using a simple optimization method, GPS-ABT is shown to converge to the optimal solution in probability, whenever the Q-value function is bounded and the gradient of the Q-value function is Lipschitz with respect to the action space. However, this method does not scale well for problems with more than 4-dimensional continuous action space. The work in \cref{sunberg2018online} are also designed for handling continuous action space problems. However, they have only been demonstrated for problems with 1-dimensional continuous action space. Another approach is to use Bayesian optimization\cref{morere2016bayesian,Mern21} for action selection, with Gaussian Process being used to represent beliefs and the estimated Q-value functions. However, they have only been demonstrated in problems with low ($< 4$) dimensional action space.

Another line of work\cref{van2011lqg} common to robotics, is to assume linear dynamics and that beliefs are Gaussian distributed. These assumptions allow one to apply LQG for solving, which has been demonstrated to show good performance on a 6-DOFs robot arm. Of course linearization does not always help. When and where linearization helps were explored in \cref{hoerger2020non}.

Another line of work\cref{Wan18:An} focuses on the problem of large discrete action space, rather than continuous action space. Problems with large discrete action space are sometimes harder than those with continuous action space because the first lack natural metric that can be used as heuristics to identify how close the performance of two actions will likely be. The work in \cref{Wan18:An} uses quantile statistics to construct a two-stage sampling mechanism for action selection, and has since been demonstrated to perform well on a logistic problem with up to 1M actions\cref{Wan19:Inventory}.

\subsection{Complex Dynamics}
\label{s:complexDyn}

To compute good approximate solutions, the above mentioned approximate POMDP solvers rely on a large number of forward simulations. They assume that each single-step forward simulation can be computed almost instantaneously. However, this assumption is false for robots with complex dynamics ---that is, robots whose dynamics are non-linear and has no closed form solution---, where a single-step forward simulation may involve solving (Partial) Differential Equation(s), which is expensive to compute. Such complex dynamics are often required when a robot needs to perform fine motion, such as, opening screws, or when a robot operates near its maximum capability, such as, car racing.  

The work in \cref{brunskill2008continuous} proposes to represent complex dynamics as switching state-space dynamics model (hybrid dynamics model), and then proposes an offline sampling-based solver for POMDPs with such a hybrid dynamics model. Another method\cref{van2012motion}, which is typical for robotics applications, is to linearize the dynamics and uses LQR, a known method from control. Whilst the work in\cref{Hoe19:Multilevel} usess MCTS-based online solvers, but apply the Multi Level Monte Carlo (MLMC) to compute the single-step forward simulations. The MLMC is used to approximate dynamic computation with varying level of fidelity, with the goal of using the expensive original dynamics only occasionally, while the majority of the approximation uses less fidelity dynamics that are less expensive to compute.

\section{Applying POMDPs to Physical Robots}
\label{s:implementation}

Given the current scalability of POMDP solving, POMDPs have been applied to solve planning and control problems in various physical robot applications, including in a robot demonstration spanning over a 7 consecutive days and 7 hours per day at SIMPAR 2018 and ICRA 2018\cref{Hoe19:POMDP}. This section discusses some of the available software and lessons learned from applying POMDPs to physical robots.

\subsection{Software}

There has been a number of software tools for solving POMDPs being released as open-source software. For instance:
\begin{itemize}
\item Symbolic Perseus\cref{Pou:Parseus} implements Perseus\cref{spaan2005perseus} but with Algebraic Decision Diagrams (ADD) for factored representation\cref{poupart2005exploiting}. It accepts a text file with SPUDD format\cref{Hoe99:Spudd} as its inputs. Symbolic Perseus is written in Java and Matlab, and requires Matlab's Java Virtual Machine. 
\item ZMDP\cref{Smi:zmdp} implements HSVI\cref{Smi04:Heuristic} and HSVI2\cref{Smi05:Point}. It is written in C++ and accepts text file with the Cassandra file format\cref{Cas17:POMDP}, which represents flat POMDP models, as inputs. 
\item Approximate POMDP Planning (APPL) Toolkit \cref{NUS:appl1} implements SARSOP\cref{Kur08:Sarsop}. It is written in C++. As its inputs, it accepts a text file in either the Cassandra \cref{Cas17:POMDP} or PomdpX file formats\cref{NUS:appl}. The latter represents factored representation and explicit separation of fully observed and partially observed state variables\cref{Ong10:Planning}.
\item APPL-online\cref{NUS:appl2} implements DESPOT\cref{somani2013despot} and is written in C++.
\item Toolkit for approximating and Adapting POMDP solutions In Realtime (TAPIR)\cref{tapir} implements ABT\cref{Kur13:An} and is written in C++. 
\item On-line POMDP Planning Toolkit (OPPT)\cref{rdl} is a software toolkit that provides a framework to ease interfacing with ROS. The POMDP model can be provided in two modes. First is via a text file where users can specify parameters for uncertainty. This mode of input is specifically designed for robot motion planning problems. For a more general problem, users can encode POMDP problems as plugins, with one plugin for each component (transition, observation, and reward functions). The default solver for this toolkit is ABT\cref{Kur13:An}, though OPPT provides interface to incorporate other solvers too. OPPT is written in C++.
\item pomdp\_py\cref{h2r} is a general purpose POMDP solving library, written in Python and Cython. It provides programming interface to implement POMDP models and solvers. 
\end{itemize}

\subsection{Implementation Tips}

\textbf{Parallelizing belief update, planning, and execution.} 
Naive implementation of POMDP solvers, and specifically online solvers described in \sref{s:pomdpSolvers},  are sequential ---that is, belief update, then computing the best action to perform from the new belief, and finally executing the action. However, such an implementation often cause delays during execution, due to the often expensive computation to update beliefs and compute the best action from  the new belief. These delays can be reduced by parallelizing the belief update and best action computation processes, and starting the computation as soon as an action started being executed\cref{Hoe19:POMDP}. \\
For instance, suppose the robot is at belief \bel and has just started execution of the action $\act^* \in \actSpace$. Then, if the belief update performs the Sequential-Importance-Resampling (SIR) particle filter, the SIR process can start as soon as the robot decides to execute $\act^*$. SIR particle filter consists of two steps. First is sampling from a proposal distribution, which in our case, $\stp \sim T(\st, \act^*, S')$ where $\st \in \stSpace$ are sampled from \bel. Second is updating the importance weights of the samples $\stp$ based on the observation $\obs \in \obsSpace$ perceived.  The first step of drawing samples can start once the robot decides to execute $\act^*$. By doing so, once the action is completely executed and an observation is perceived, SIR only needs to update the importance weights, which can be done fast. \\
In computing the best action, if ABT is used, one can sample additional episodes, starting from states  sampled from the current belief \bel and performing $\act^*$, as soon as the robot decides to execute $\act^*$, so as to improve the policy within the entire descendent of $\bel$ via $\act^*$ in the belief tree \belTree. This strategy increases the chances that after $\act^*$ is completely executed and the belief is updated based on the observation perceived, a good policy for the next belief is readily available in \belTree. Of course, there are cases where even with the above strategy, the robot perceives observations that were not explored in \belTree. In such cases, one can reuse value estimates from nearby beliefs or restart planning from scratch.

\textbf{Distance function.} 
Some of the solvers use distance function between beliefs as part of their computation, often as a heuristic to assess how close two beliefs are. For this purpose, L1 metric and KL-divergence have often been used. However, for robotics problems, it is often desirable to account for the state space distance when computing distance between beliefs. For this purpose, Earth Mover Distance (EMD) is often more suitable than L1 or KL-divergence\cref{Lit15:The}.  Fast EMD computation has been developed in the computer vision community, including incorporated in OpenCV. 

\subsection{Some Notes on the POMDP Models}
\label{s:noteModels}

Below are three main concerns one often have about using POMDPs, together with a discussion that we hope would reduce such concerns. 

\textbf{How difficult is it to generate a suitable POMDP model?} 
A POMDP model consist of six components. The state, action, and observation spaces are generally easy to define. However, the transition, observation, and reward functions are indeed harder to define. To model the transition and observation functions, one can use information about potential errors and develop a relatively conservative estimate, learn from data, or a combination of both. Setting the reward function can be quite involved if one wants to use the reward function as a heuristics to help guide the search. However, if we set the reward function to reflect desirability of being in a state, rather than as heuristics, then setting the reward functions are easier, though in this case, we do need solvers with good scalability. \\
Moreover, in most robotics problems, one can generate good POMDP policies without accurate POMDP models. For instance, the transition and observation functions used in a POMDP demonstration at ICRA'18\cref{Hoe19:POMDP} are a very rough estimate, learned using a simple likelihood approach from a small amount of data. However, the POMDP strategies generated for this rough model resulted in a 100\% success rate, while those generated without consideration of uncertainty resulted in only 35\% success rate\cref{Hoe19:POMDP}. 
Furthermore, results on  end-to-end POMDP model learning and solving\cref{karkus2017qmdpnet} indicate the models learned can often be different from the correct model but the policy generated are performing well.

\textbf{What do we really gain by formulating and solving a problem as a POMDP?} 
The short answer is robustness. A more nuanced answer is that by constructing a feedback policy that quantifies uncertainty, POMDPs can automatically balance trade-offs between information gathering actions and  performing actions to achieve its task. In fact, it can even identify actions that could achieve both, as highlighted in the simulation result of\cref{Kur11:Motion}. Such a capability is important when the solution space is small, such as when robots must operate in cluttered or confined environment.

\textbf{Isn't first order Markov too restrictive?} 
One concern with POMDPs is the first order Markov requirement. It might be useful to clarify that the POMDP policy actually accounts for the entire history because a POMDP policy maps beliefs to actions, and beliefs are sufficient statistics of the entire history\cref{hauskrecht2000value}. The first order Markov is indeed required for the transition dynamics and observation functions. However, the first order Markov requirements for those functions are also common in state space control\cref{friedland2012control}, which is often used in robotics.

\section{Discussion}

The above two sections present an overview of some of the general sampling-based approximate POMDP solvers. This is by no means an exhaustive list of POMDP solving approaches. For instance, we did not cover policy search based approaches that have been introduced since\cref{Ng00Pegasus}. We also did not provide coverage on work that restricts beliefs to be Gaussian and those that extend deterministic sampling-based motion planning, such as the Probabilistic Roadmap\cref{kavraki1996probabilistic}, to belief space planning  beyond the few mentioned above. Furthermore, the surveys \cref{shani2013survey} and \cref{ross2008online} provide  a more exhaustive list on offline sampling-based approximate POMDP solvers up to 2013 and approximate online POMDP solvers up to 2008, respectively. However, we focus to elaborate computational issues that have hindered the practicality of POMDPs in robotics and  elucidate ideas that have alleviated them. 

\subsection{Comparison to Sampling-Based Motion Planning}

Taking a step back, it is interesting to note that the key techniques and progressions that enable POMDPs to become practical in robotics is close to those of motion planning. \tref{t:motion-pomdp} tries to capture this similarity. 

\begin{table}[!htbp]
\caption{Sampling-based POMDP Solvers and Motion Planning}\vspace{12pt}
\label{t:motion-pomdp}
\centering
\begin{tabular}{|p{4.5cm}|p{6cm}|p{6cm}|}
\hline 
\multirow{2}{*}{Components} & \multicolumn{2}{|c|}{Towards Sampling-based methods} \\
& POMDPs & Motion Planning \\
\hline\hline
Helpful concepts and theoretical results & Optimal value function is (or can be approximated arbitrarily close to) a piecewise linear convex function\cref{sondik1971optimal} for \alp-vectors representation, and heuristic based forward search\cref{bonet1998solving} for online tree-based solvers. 
& The  configuration space\cref{lozano1979algorithm}. \\
\hline
Fast primitive computation & Point-based dynamic programming\cref{zhang2001speeding} for offline solvers and Upper Confidence Tree\cref{kocsis2006bandit} for online solvers. & Fast collision-check \cref{larsen1999fast, gottschalk1996obbtree}. \\
\hline
Combined sampling-based with a more classical approach & Combined point-based and standard dynamic programming backup\cref{zhang2001speeding}. & Potential field with randomization to exit local minima\cref{barraquand1990monte} and the Ariadne's Clew algorithm\cref{bessiere1993ariadne}. \\
\hline
Full sampling-based to solve the problem start to become scalable & Off-line, where a good policy \pol is computed prior to execution, and during execution, the action $\pol(\bel)$ will be executed whenever the agent is at belief \bel, started with \cref{Pin03:Point}. & Multi-query, where the goal is to construct a compact representation of the free space component of the robot's configuration space, started with \cref{kavraki1996probabilistic}. \\
\hline
Sampling-based to solve a smaller problem (potentially, iteratively) to improve scalability & On-line, where the best action to perform is computed only for the belief at the current time-step, popularized by\cref{silver2010monte}. & Single-query, where  the goal is to answer a query to move the robot from a given initial to goal configurations, started with \cref{hsu1997path}.\\
\hline
\end{tabular}
\end{table}

\subsection{Relation to Learning}
 
The POMDP is closely related to learning. It is a basic representation for model-based Bayesian Reinforcement Learning\cref{Gha15:Bayesian}. Reinforcement Learning (RL) can be defined as a Markov Decision Process (MDP, the fully observed version of POMDP) with missing components. Since MDP models a fully observed system, MDP is defined as \mdpTuple, where \mdpStSpace is the state space, \mdpActSpace is the action space, $\mdpTrans(\st, \act, \stp)$ is the transition function, representing the conditional probability function of moving to state $\stp \in \mdpStSpace$ after performing action $\act \in \mdpActSpace$ from state $\st \in \mdpStSpace$, and \mdpRew is the reward function. RL is then defined as MDP with initially unknown transition and/or reward functions. 

POMDP representation of the RL problem is to model uncertainty over the \mdpTrans and \mdpRew as probability distribution functions, generally as parametric distribution functions. The parameters of these functions are then set as partially observed state variables of the POMDP agent. As the POMDP agent perceives observations, its understanding about the true parameters improve. By solving this POMDP problem, the agent automatically balances the trade-off between information gathering actions to reduce uncertainty on the parameters and actions to achieve the task, and identifies actions that help improve both model understanding and task attainment. The difficulty of this approach of RL is that naive modelling often results in POMDP models that are much larger than the size of problems that state of the art POMDP solvers can handle. 

On another note, as mentioned in \sref{s:noteModels}, the transition and observation functions of POMDPs have often been learned from data. More recently, deep learning has been applied to solve POMDPs when its model is not fully known. Some of the early work\cref{deep_learning_pomdp, deep_learning_2,dl_pomdp_2} are model free, they directly learn the policy or value function without learning the POMDP model. However, better generalization has been achieved with methods\cref{rcnn_qmdp, karkus2017qmdpnet, Oh17:Value, Igl18:Deep, Col20:Locally} that embed the POMDP structure inside a neural network and training the network to learn a policy or value function, thereby combining model-based and model-free methods.

\section{Conclusion}

The Partially Observable Markov Decision Process (POMDP) is a mathematical framework for planning under uncertainty, and specifically for non-deterministic and partially observable scenarios. Finding the optimal solution to a POMDP problem is computationally intractable. However, sampling-based methods are now available to compute good optimal solutions ---ones that significantly improve the robustness of robotics systems--- within reasonable computational resources. Improving the scalability of POMDPs from a mere theoretical concept that can only work for small toy problems into a software tool that can be applied to a variety of realistic robotics problems requires multiple issues to be overcome. For robotics problems, five major issues are large state, observation, and actions spaces, long planning horizon, and complex dynamics. Various sampling-based techniques have been proposed to alleviate these issues. Software implementations of some of these methods and interfaces to typical robotics software are now available as Open Source Software to ease applying POMDPs to robotics problems. This paper presents an overview of the issues, methods, and practical tips on applying POMDPs to robotics. 

Although some scalability issues in POMDPs remain, existing methods are efficient enough to improve the robustness of many robotics problems. We hope this paper could provide insights on the current state of POMDPs and help bring more awareness on the practicality of POMDPs in robotics.


\section*{ACKNOWLEDGMENTS}
This work is supported by the ANU Futures Scheme.

%

\bibliographystyle{abbrv}
\bibliography{refs}

%
%
%

\end{document}